\documentclass[11pt]{article}

\usepackage{preprint}

\usepackage{times}
\usepackage{latexsym}
\usepackage{floatflt}
\usepackage[T1]{fontenc}

\usepackage[utf8]{inputenc}

\usepackage{microtype}

\usepackage{inconsolata}

\usepackage{graphicx}
\usepackage[utf8]{inputenc} 
\usepackage[T1]{fontenc}    
\usepackage{hyperref}       
\usepackage{url}            
\usepackage{booktabs}       
\usepackage{amsfonts}       
\usepackage{nicefrac}       
\usepackage{xspace}
\usepackage{microtype}      
\usepackage{xcolor}         
\usepackage{graphicx}
\usepackage{adjustbox}
\usepackage{textcomp}
\usepackage{listings}
\usepackage{subcaption}
\usepackage{amsmath}
\usepackage{xcolor}
\usepackage{wrapfig}
\usepackage{tabularray}
\usepackage{cleveref}
\usepackage{multirow}
\usepackage{wasysym}

\newcommand{\fullcirc}{\CIRCLE}
\newcommand{\halfcirc}{\LEFTcircle}
\newcommand{\emptycirc}{\Circle}
\newenvironment{btHighlight}[1][]
{\begingroup\tikzset{bt@Highlight@par/.style={#1}}\begin{lrbox}{\@tempboxa}}
{\end{lrbox}\bt@HL@box[bt@Highlight@par]{\@tempboxa}\endgroup}

\newcommand\btHL[1][]{%
  \begin{btHighlight}[#1]\bgroup\aftergroup\bt@HL@endenv%
}
\def\bt@HL@endenv{%
  \end{btHighlight}%
  \egroup
}
\lstdefinestyle{XXX}{
basicstyle=\ttfamily\footnotesize,
moredelim=**[is][\btHL]{!}{!},
escapeinside={&}{&},
keywordstyle=\color{blue}\bfseries,
commentstyle=\color{gray}\itshape,
stringstyle=\color{teal},
}
\lstdefinestyle{sty2}{
    backgroundcolor=\color{gray!10},   
    commentstyle=\color{green!50!black},
    keywordstyle=\color{blue}\bfseries,
    numberstyle=\tiny\color{gray},
    stringstyle=\color{orange},
    basicstyle=\ttfamily\small,
    breaklines=true,                 
    captionpos=b,                    
    keepspaces=true,                 
    numbers=left,                    
    numbersep=5pt,                  
    showspaces=false,                
    showstringspaces=false,
    showtabs=false,                  
    tabsize=4,
    frame=single,
    rulecolor=\color{black},
}

\newcommand{\name}{AgentFly\xspace}
\lstdefinestyle{paper}{
  language=Python, basicstyle=\ttfamily\small, columns=fullflexible, keepspaces=true,
  keywordstyle=\bfseries, commentstyle=\color{black!55}, stringstyle=\color{black}, upquote=true,
  alsoletter={@}, keywords={def,return,async,await,class,import,from,@tool,@reward}, keywords=[2]{},
  numbers=left, numberstyle=\scriptsize\color{black!45}, numbersep=8pt,
  frame=tb, framerule=0.6pt, aboveskip=6pt, belowskip=4pt,
  xleftmargin=0.17\linewidth, xrightmargin=0.17\linewidth, showstringspaces=false,
  captionpos=b}
\lstdefinestyle{paperhalf}{style=paper, basicstyle=\ttfamily\footnotesize,
  xleftmargin=1.6em, xrightmargin=0pt, numbersep=6pt, aboveskip=2pt, belowskip=2pt}
\usepackage{fancyvrb}
\definecolor{codeKW}{HTML}{AF5F00}\definecolor{codeFN}{HTML}{7B3FA0}
\definecolor{codeCM}{HTML}{6B7280}\definecolor{codeST}{HTML}{1F6F43}
\definecolor{hlRes}{HTML}{FFE0B2}\definecolor{hlCtx}{HTML}{CDEBC5}
\newcommand{\ckw}[1]{\textcolor{codeKW}{#1}}
\newcommand{\cfn}[1]{\textcolor{codeFN}{#1}}
\newcommand{\ccm}[1]{\textcolor{codeCM}{#1}}
\newcommand{\cst}[1]{\textcolor{codeST}{#1}}
\newcommand{\hlres}[1]{{\setlength{\fboxsep}{0pt}\colorbox{hlRes}{\vphantom{Ag(}#1}}}
\newcommand{\hlctx}[1]{{\setlength{\fboxsep}{0pt}\colorbox{hlCtx}{\vphantom{Ag(}#1}}}
\newtcolorbox{codebox}{sharp corners, colback=white, colframe=black!85, boxrule=0.4pt,
  left=3pt, right=2pt, top=1pt, bottom=1pt, equal height group=codeboxes}
\crefname{lstlisting}{Listing}{Listings}
\Crefname{lstlisting}{Listing}{Listings}
\usetikzlibrary{positioning,fit,calc,arrows.meta,backgrounds}

\definecolor{genfill}{HTML}{EDE9FE}\definecolor{gendraw}{HTML}{6D28D9}
\definecolor{toolfill}{HTML}{E0F2FE}\definecolor{tooldraw}{HTML}{0284C7}
\definecolor{rewfill}{HTML}{FEF3C7}\definecolor{rewdraw}{HTML}{D97706}
\definecolor{resfill}{HTML}{DCFCE7}\definecolor{resdraw}{HTML}{16A34A}
\definecolor{ctxfill}{HTML}{F3F4F6}\definecolor{ctxdraw}{HTML}{6B7280}
\definecolor{layerfill}{HTML}{FAFAFB}\definecolor{layerdraw}{HTML}{D1D5DB}
\definecolor{codekw}{HTML}{6D28D9}\definecolor{codefn}{HTML}{0369A1}
\definecolor{codestr}{HTML}{047857}\definecolor{codearg}{HTML}{B45309}

\tikzset{
  af/.style={font=\sffamily\footnotesize, >={Stealth[length=4pt]}},
  chip/.style={rounded corners=2pt, minimum height=12pt, inner xsep=4pt, inner ysep=2pt, text height=1.5ex, text depth=0.3ex,
               font=\ttfamily\scriptsize, line width=0.5pt},
  gen/.style={chip, fill=genfill, draw=gendraw},
  tool/.style={chip, fill=toolfill, draw=tooldraw},
  rew/.style={chip, fill=rewfill, draw=rewdraw},
  res/.style={chip, fill=resfill, draw=resdraw},
  ctx/.style={chip, fill=ctxfill, draw=ctxdraw},
  layer/.style={rounded corners=4pt, fill=layerfill, draw=layerdraw, line width=0.5pt},
  panel/.style={rounded corners=3pt, fill=white, draw=layerdraw, line width=0.5pt},
  layerlabel/.style={font=\sffamily\bfseries\small, text=afdeep, anchor=west},
  link/.style={draw=ctxdraw, line width=0.5pt, densely dotted},
  flow/.style={draw=afink, line width=0.7pt, ->},
  pill/.style={rounded corners=5pt, minimum height=10pt, inner xsep=4pt, inner ysep=1pt,
               font=\sffamily\scriptsize},
}

\newcommand{\tool}[1]{\texttt{#1}\xspace}

\title{\name: Scaling Agentic Reinforcement Learning\\with Unified Resource System}
\toplogos{\includegraphics[height=32pt]{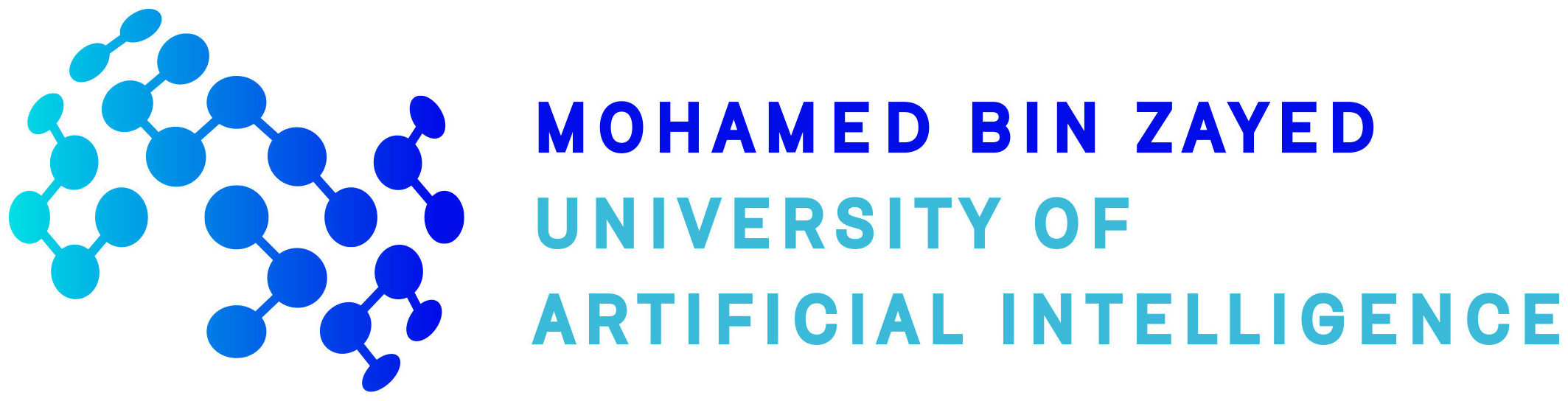}}

\author{\textbf{Renxi Wang}$^{1}$\quad \textbf{Bilal Elbouardi}$^{1}$\quad \textbf{Rifo Ahmad Genadi}$^{1}$\quad \textbf{Yongxin Wang}$^{1}$\quad \textbf{Shu Wang}$^{2}$\\[2pt]
\textbf{Fajri Koto}$^{1}$\quad \textbf{Zhengzhong Liu}$^{1}$\quad \textbf{Timothy Baldwin}$^{1}$\quad \textbf{Haonan Li}$^{1}$}

\affiliations{$^1$Mohamed bin Zayed University of Artificial Intelligence \quad $^2$University of California, Los Angeles\\[2pt]{\ttfamily\{renxi.wang,haonan.li\}@mbzuai.ac.ae}\\[2pt]\url{https://github.com/Agent-One-Lab/AgentFly}}

\begin{document}
\maketitle
\begin{abstract}
Methods to build LLM agents have evolved from prompt engineering and supervised finetuning to agentic reinforcement learning (agentic RL). However, agentic RL remains bottlenecked by its surrounding systems: agents must interact with heterogeneous environments, such as sandboxes, model services, and external APIs. Their allocation, reuse, and lifecycle dominate rollout cost and cap the scale at which training becomes practical. In this work, we present \name, an agentic RL framework built with a unified resource layer that treats each of these environments as a distinct, typed resource scheduled through one engine, with per-tool acquisition for multi-turn reuse, asynchronous backpressure, and rollout versus global-scoped lifecycles. \name adopts a four-layer design: (I) agent layer that abstracts the agent, tool, and reward concepts, decomposing agentic RL into defining agents, tools, and reward functions; (II) rollout layer that composes these into agent loops and computes rewards; (III) context layer that organizes rollouts, injects contextual information, and arranges resources; and (IV) a low-level resource layer that performs resource management. We provide a suite of prebuilt tools and environments, demonstrate successful agent training across multiple tasks and models, and report the first controlled cross-framework throughput comparison against agentic RL frameworks.
\end{abstract}

\section{Introduction}

Large language model (LLM) agents have been widely used in research and applications \citep{wuautogen,yao2023react,sumers2023cognitive}, where an LLM is taken as the core to interact with tools, environments, and APIs \citep{qintoolllm,wangtoolgen}. Equipped with these external interfaces, LLMs can complete tasks beyond their intrinsic capability, such as fetching online information or executing code. Early agents are built with prompt engineering, where all task information and available actions are given in the prompt \citep{wuautogen}. Training-based methods such as supervised finetuning further improve LLMs by learning from agent interaction trajectories \citep{zeng-etal-2024-agenttuning,chen2023fireact}. Reinforcement learning (RL) offers a more powerful paradigm, as it directly optimizes long-horizon objectives and has been shown to substantially enhance the reasoning, coding, and other capabilities of LLMs. However, applying RL to agentic settings has been shown to be challenging due to several issues.

\begin{figure*}[t]
  \IfFileExists{images/overview_drawio.pdf}{\includegraphics[width=\linewidth]{images/overview_drawio.pdf}}{\resizebox{\linewidth}{!}{
\begin{tikzpicture}[x=1cm, y=1cm, font=\sffamily\footnotesize, >={Stealth[length=4pt]}]
\definecolor{archline}{HTML}{404040}\definecolor{archgray}{HTML}{666666}
\definecolor{agF}{HTML}{FFF2CC}\definecolor{agD}{HTML}{D6B656}
\definecolor{roF}{HTML}{C9DAF8}\definecolor{roD}{HTML}{6C8EBF}
\definecolor{ctF}{HTML}{D9EAD3}\definecolor{ctD}{HTML}{82B366}
\definecolor{reF}{HTML}{FCE5CD}\definecolor{reD}{HTML}{D79B00}
\definecolor{trF}{HTML}{EFEFEF}\definecolor{trD}{HTML}{999999}
\tikzset{
  lay/.style={draw=archline, line width=0.6pt, fill=white},
  laylab/.style={font=\sffamily\bfseries\small, rotate=90, align=center},
  comp/.style={draw=archline, line width=0.5pt, fill=white, align=center,
               minimum height=0.8cm, inner sep=2pt},
  acomp/.style={comp},
  cAg/.style={fill=agF, draw=agD}, cRo/.style={fill=roF, draw=roD}, cCt/.style={fill=ctF, draw=ctD},
  cRe/.style={fill=reF, draw=reD}, cTr/.style={fill=trF, draw=trD},
  lstep/.style={draw=archline, line width=0.5pt, fill=white, minimum height=0.55cm,
               minimum width=1.75cm, inner sep=2pt},
  op/.style={font=\ttfamily\scriptsize, fill=white, inner sep=1.5pt},
  arr/.style={->, draw=archline, line width=0.6pt},
  sub/.style={font=\sffamily\scriptsize, text=archgray},
}
\newcommand{\cn}[2]{{\bfseries #1}\\[-1pt]{\scriptsize\color{archgray}#2}}
\def\W{12.8}

\foreach \top/\bot/\name/\f/\c in {0/-1.3/Agent/agF/agD, -1.75/-3.2/Rollout/roF/roD,
                                    -3.65/-4.95/Context/ctF/ctD, -5.4/-7.6/Resource/reF/reD}{
  \draw[lay, fill=white, draw=black, dashed] (0,\top) rectangle (\W,\bot);
  \draw[draw=black, dashed, line width=0.6pt] (0.55,\top) -- (0.55,\bot);
  \node[laylab] at (0.275,{(\top+\bot)/2}) {\name};
}

\node[comp, cAg, minimum width=3.8cm] (ag) at (2.7,-0.65) {\cn{Agent}{model, prompt, parser, memory}};
\node[comp, cAg, minimum width=3.8cm] (tl) at (6.7,-0.65) {\cn{Tools}{\texttt{@tool} functions}};
\node[comp, cAg, minimum width=3.8cm] (rw) at (10.7,-0.65) {\cn{Reward}{\texttt{@reward} function}};

\node[lstep, cRo] (g) at (1.8,-2.3) {generate};
\node[lstep, cRo] (t) at (4.05,-2.3) {tool call};
\node[lstep, cRo] (o) at (6.3,-2.3) {observation};
\node (dots) at (7.8,-2.3) {\dots};
\node[lstep, cRo] (r) at (9.3,-2.3) {reward};
\draw[arr] (g) -- (t); \draw[arr] (t) -- (o); \draw[arr] (o) -- (dots); \draw[arr] (dots) -- (r);
\draw[arr] (o.south) to[out=-155, in=-25] node[sub, above=0pt, pos=0.5] {multi-turn} (g.south);
\node[sub, align=right, anchor=east] at (12.65,-2.3) {$N$ chains\\in parallel, async};

\node[comp, cCt, minimum width=5.4cm] (inj) at (3.5,-4.3) {\cn{Information injection}{rollout id, input data, reference answers}};
\node[acomp, cCt, minimum width=6.2cm] (brk) at (9.5,-4.3) {\cn{Resource broker}{tracks resources per rollout; releases at end}};

\node[sub, anchor=west] at (0.75,-5.72) {one pool per typed spec \textperiodcentered{} capacity limit with async wait \textperiodcentered{} rollout or global scope};
\foreach \n/\x in {container/2.2, vLLM server/5.2, model API/8.2, in-process env/11.2}
  \node[acomp, cRe, minimum width=2.8cm, minimum height=0.6cm] at (\x,-6.3) {\n};
\foreach \n/\x in {Local runner/2.2, Slurm runner/5.2, Ray runner/8.2}
  \node[comp, cRe, minimum width=2.8cm, minimum height=0.5cm, font=\sffamily\scriptsize] at (\x,-7.1) {\n};
\node[op, fill=none] at (11.2,-7.1) {start/reset/release};

\draw[arr] (6.7,-1.3) -- node[op, right=1pt] {agent, tools, reward} (6.7,-1.75);
\draw[arr] (1.8,-3.2) -- node[op, right=1pt] {register / end rollout} (1.8,-3.65);
\draw[arr] (8.6,-3.2) -- node[op, left=1pt] {acquire(spec, id)} (8.6,-3.65);
\draw[arr] (10.4,-3.65) -- node[op, right=1pt] {instance} (10.4,-3.2);
\draw[arr] (8.6,-4.95) -- node[op, left=1pt] {acquire / release} (8.6,-5.4);
\draw[arr] (10.4,-5.4) -- node[op, right=1pt] {instance} (10.4,-4.95);

\def\TL{14.35}\def\TR{17.25}
\draw[lay, fill=white, draw=black, dashed] (\TL,0) rectangle (\TR,-7.6);
\draw[draw=black, dashed, line width=0.6pt] (\TL+0.55,0) -- (\TL+0.55,-7.6);
\node[laylab] at (\TL+0.275,-3.8) {RL trainer (verl)};
\node[comp, cTr, minimum width=1.9cm, minimum height=0.6cm] (w) at (15.97,-0.65) {policy weights};
\node[comp, cTr, minimum width=1.9cm, minimum height=0.6cm] (lp) at (15.97,-2.3) {log-probs};
\node[comp, cTr, minimum width=1.9cm, minimum height=0.6cm] (ad) at (15.97,-4.3) {advantages};
\node[comp, cTr, minimum width=1.9cm, minimum height=0.6cm] (up) at (15.97,-6.3) {actor update};
\draw[arr] (lp) -- (ad); \draw[arr] (ad) -- (up);
\draw[arr] (up.east) -- ++(0.15,0) |- (w.east);
\draw[arr] (\W,-2.3) -- node[sub, above=1pt, text=archline] {trajectories} (\TL,-2.3);
\draw[arr] (\TL,-0.65) -- node[sub, above=1pt, text=archline] {weights} (\W,-0.65);
\end{tikzpicture}}}
  \caption{Architecture of \name. The agent layer holds the user-defined agent, tools, and reward; the rollout layer runs the multi-turn agent loop for many chains in parallel; the context layer injects rollout information and brokers resources; and the resource layer manages one pool per typed resource specification, launched by pluggable runners. Together, the context and resource layers form the unified resource system. Completed rollouts are sent to the RL trainer (\texttt{verl}) as trajectories, and updated weights are returned to the agent.}
  \label{fig:overview}
\end{figure*}

Agentic RL involves multi-turn interactions where LLM-generated actions alternate with environment observations \citep{yao2023react}. We leave optimizing long-horizon trajectories to existing RL trainers, while aiming at the environment bottleneck: For LLM agents, rollout repeatedly allocates, drives, and tears down external environments, like sandboxes, model services, and external APIs; reuse and lifecycle costs can even exceed text generation. These environments are heterogeneous in kind and lifetime, and managing them concurrently at high throughput, without starving the trainer, is the central engineering bottleneck.

We explore the framework designs that manage these heterogeneous resources under a single abstraction, while balancing modularity and RL training efficiency, mitigating the gap in building systems that are both developer-friendly and scalable.

We propose \name, a framework for training LLM agents with reinforcement learning. Our framework uses \texttt{verl} \citep{sheng2024hybridflow} as the RL trainer while building support for a more modular and scalable agentic RL setup. We made the following efforts to mitigate the above gaps:
\begin{itemize}[nosep]
  \item \textbf{Unified resource system.} We introduce a \emph{unified resource system} that schedules containers, local model services, and external model APIs as distinct, typed resources, with per-tool acquisition for multi-turn reuse, asynchronous backpressure, and managed lifecycles.
  \item \textbf{Four-layer design.} We adopt a four-layer design that decomposes agentic RL into defining agents, tools, and reward functions, while the lower layers handle rollout composition, context organization, and low-level resource management. Users can focus on agent workflows without managing training or resource-allocation details.
  \item \textbf{Broad validation.} We pre-build a suite of environments and tools and validate \name on six representative tasks that cover various interaction patterns and reasoning requirements. These experiments demonstrate the flexibility, scalability, and effectiveness of our framework in supporting diverse agentic RL training scenarios.
  \item \textbf{Cross-framework throughput.} We compare with multiple existing agentic RL frameworks in heavy agentic environment settings, where \name shows significantly higher efficiency and scalability.
\end{itemize}

\section{\name}

We first introduce the high-level design of the framework, then describe the details of the whole system, and finally show the supported tools and tasks.

\subsection{High-Level Design Choices}

\begin{figure}[t]
\centering
\begin{minipage}[t]{0.485\linewidth}
\centering
\resizebox{\linewidth}{!}{
\begin{tikzpicture}[x=1cm, y=1cm, font=\sffamily\footnotesize, >={Stealth[length=4pt]}]
\definecolor{archgray}{HTML}{666666}
\definecolor{roF}{HTML}{C9DAF8}\definecolor{roD}{HTML}{6C8EBF}
\definecolor{reF}{HTML}{FCE5CD}\definecolor{reD}{HTML}{D79B00}
\definecolor{trF}{HTML}{EFEFEF}\definecolor{trD}{HTML}{999999}
\tikzset{
  lay/.style={draw=black, dashed, line width=0.6pt, fill=white},
  laylab/.style={font=\sffamily\bfseries\small, rotate=90, align=center},
  comp/.style={draw=roD, line width=0.5pt, fill=roF, minimum height=0.5cm,
               inner sep=2pt, anchor=west, font=\ttfamily\scriptsize},
  rcomp/.style={comp, draw=reD, fill=reF},
  tag/.style={draw=trD, fill=trF, line width=0.4pt, font=\sffamily\scriptsize,
              inner xsep=3pt, inner ysep=1pt},
  sub/.style={font=\sffamily\scriptsize, text=archgray},
}
\draw[lay] (0,0) rectangle (8.2,-4.32);
\draw[black, dashed, line width=0.6pt] (0.55,0) -- (0.55,-4.32);
\node[laylab] at (0.275,-2.16) {Agent.rollout};

\draw[archgray, densely dotted, line width=0.6pt] (0.75,-0.2) rectangle (8.0,-2.3);
\node[sub, anchor=north west] at (0.8,-0.22) {agent loop, per turn};
\node[comp] at (0.95,-1.05) {Model.generate};
\node[comp] (tc) at (0.95,-1.75) {Tool.call};
\node[rcomp] at ($(tc.east)+(0.15,0)$) {Resource.acquire/execute};

\node[comp] (rc) at (0.95,-2.95) {Reward.call};
\node[rcomp] at ($(rc.east)+(0.15,0)$) {Resource.acquire/execute};
\node[rcomp] at (0.95,-3.7) {Resource.release};

\foreach \y in {-1.05,-1.75,-2.95,-3.7} \node[tag, anchor=east] at (7.9,\y) {async};
\end{tikzpicture}}
\end{minipage}\hfill
\begin{minipage}[t]{0.485\linewidth}
\centering
\resizebox{\linewidth}{!}{
\begin{tikzpicture}[x=1cm, y=1cm, font=\sffamily\footnotesize]
\definecolor{archline}{HTML}{404040}\definecolor{archgray}{HTML}{666666}
\definecolor{roF}{HTML}{C9DAF8}\definecolor{roD}{HTML}{6C8EBF}\definecolor{reF}{HTML}{FCE5CD}\definecolor{reD}{HTML}{D79B00}
\definecolor{agF}{HTML}{FFF2CC}\definecolor{agD}{HTML}{D6B656}
\tikzset{seg/.style={line width=0.45pt},
  sgen/.style={seg, draw=roD, fill=roF}, stool/.style={seg, draw=reD, fill=reF},
  srew/.style={seg, draw=agD, fill=agF},
  idle/.style={draw=archgray, densely dotted, line width=0.4pt},
  lab/.style={font=\sffamily\scriptsize, text=archgray}, ttl/.style={font=\sffamily\bfseries\scriptsize, anchor=west}}
\draw[draw=black, dashed, line width=0.6pt, fill=white] (0,0) rectangle (8.2,-4.32);
\draw[black, dashed, line width=0.6pt] (0.55,0) -- (0.55,-4.32);
\node[font=\sffamily\bfseries\small, rotate=90] at (0.275,-2.16) {Rollout schedule};
\node[ttl] at (0.7,-0.22) {Batch-synchronous};
\node[lab, anchor=east] at (1.17,-0.600) {R1};
\draw[sgen] (1.275,-0.450) rectangle (2.369,-0.750);
\draw[idle] (2.394,-0.600) -- (2.623,-0.600);
\draw[stool] (2.648,-0.450) rectangle (3.284,-0.750);
\draw[idle] (3.309,-0.600) -- (4.339,-0.600);
\draw[sgen] (4.364,-0.450) rectangle (5.229,-0.750);
\draw[idle] (5.254,-0.600) -- (5.483,-0.600);
\draw[stool] (5.508,-0.450) rectangle (7.517,-0.750);
\draw[srew] (7.567,-0.450) rectangle (7.975,-0.750);
\node[lab, anchor=east] at (1.17,-1.000) {R2};
\draw[sgen] (1.275,-0.850) rectangle (2.026,-1.150);
\draw[idle] (2.051,-1.000) -- (2.623,-1.000);
\draw[stool] (2.648,-0.850) rectangle (4.314,-1.150);
\draw[sgen] (4.364,-0.850) rectangle (5.458,-1.150);
\draw[stool] (5.508,-0.850) rectangle (6.030,-1.150);
\draw[idle] (6.055,-1.000) -- (7.542,-1.000);
\draw[srew] (7.567,-0.850) rectangle (7.975,-1.150);
\node[lab, anchor=east] at (1.17,-1.400) {R3};
\draw[sgen] (1.275,-1.250) rectangle (2.598,-1.550);
\draw[stool] (2.648,-1.250) rectangle (3.056,-1.550);
\draw[idle] (3.081,-1.400) -- (4.339,-1.400);
\draw[sgen] (4.364,-1.250) rectangle (5.000,-1.550);
\draw[idle] (5.025,-1.400) -- (5.483,-1.400);
\draw[stool] (5.508,-1.250) rectangle (6.259,-1.550);
\draw[idle] (6.284,-1.400) -- (7.542,-1.400);
\draw[srew] (7.567,-1.250) rectangle (7.975,-1.550);
\draw[archline, dashed, line width=0.5pt] (8.000,-0.400) -- (8.000,-1.600);
\node[lab, anchor=south east] at (7.960,-0.420) {finish};
\node[ttl] at (0.7,-1.85) {\name{} (asynchronous)};
\node[lab, anchor=east] at (1.17,-2.230) {R1};
\draw[sgen] (1.275,-2.080) rectangle (2.369,-2.380);
\draw[stool] (2.419,-2.080) rectangle (3.056,-2.380);
\draw[sgen] (3.106,-2.080) rectangle (3.971,-2.380);
\draw[stool] (4.021,-2.080) rectangle (6.030,-2.380);
\draw[srew] (6.080,-2.080) rectangle (6.488,-2.380);
\node[lab, anchor=east] at (1.17,-2.630) {R2};
\draw[sgen] (1.275,-2.480) rectangle (2.026,-2.780);
\draw[stool] (2.076,-2.480) rectangle (3.742,-2.780);
\draw[sgen] (3.792,-2.480) rectangle (4.886,-2.780);
\draw[stool] (4.936,-2.480) rectangle (5.458,-2.780);
\draw[srew] (5.508,-2.480) rectangle (5.916,-2.780);
\node[lab, anchor=east] at (1.17,-3.030) {R3};
\draw[sgen] (1.275,-2.880) rectangle (2.598,-3.180);
\draw[stool] (2.648,-2.880) rectangle (3.056,-3.180);
\draw[sgen] (3.106,-2.880) rectangle (3.742,-3.180);
\draw[stool] (3.792,-2.880) rectangle (4.543,-3.180);
\draw[srew] (4.593,-2.880) rectangle (5.000,-3.180);
\draw[archline, dashed, line width=0.5pt] (6.513,-2.030) -- (6.513,-3.230);
\node[lab, anchor=south east] at (6.473,-2.050) {finish};
\draw[<->, black, line width=0.5pt] (6.513,-3.200) -- (8.000,-3.200);
\node[font=\sffamily\scriptsize, anchor=south] at (7.256,-3.200) {time saved};
\draw[->, archline, line width=0.5pt] (1.25,-3.45) -- (8.0,-3.45) node[lab, anchor=north east] at (8.0,-3.47) {time};
\draw[sgen] (1.25,-3.93) rectangle (1.60,-4.13);
\node[lab, anchor=west] at (1.65,-4.03) {generate};
\draw[stool] (2.90,-3.93) rectangle (3.25,-4.13);
\node[lab, anchor=west] at (3.30,-4.03) {tool call};
\draw[srew] (4.55,-3.93) rectangle (4.90,-4.13);
\node[lab, anchor=west] at (4.95,-4.03) {reward};
\draw[idle] (5.95,-4.03) -- (6.30,-4.03);
\node[lab, anchor=west] at (6.35,-4.03) {idle};
\end{tikzpicture}}
\end{minipage}
\caption{Fully asynchronous rollout in \name. \emph{Left:} within a rollout, generation, tool calls, reward computation, and resource operations (peach) are each awaited asynchronously, so a slow step in one rollout does not block the others. \emph{Right:} schematic schedule of three rollouts with two turns each: in a batch-synchronous design every phase waits for the slowest rollout (dotted), whereas in \name each rollout proceeds independently and the batch finishes earlier. Durations are illustrative.}
\label{fig:async}
\end{figure}

\paragraph{Decoupled Agent Module}
Reinforcement learning for LLMs already involves complex distributed systems for generation and optimization. We therefore decouple the agent module from RL training: the training process is delegated to \texttt{verl}\footnote{\url{https://github.com/volcengine/verl}}, while the agentic workflow is handled by a four-layer system that controls rollouts and returns trajectories to the trainer. The two sides communicate only through trajectories, so the agent module evolves independently of the training infrastructure. \Cref{fig:overview} illustrates the overall structure of the framework, including this decoupling.

\paragraph{Tool Abstraction}
We use \texttt{tool} as a unified abstraction for all external interfaces that LLM agents interact with, including functions, APIs, and environments. Every agent interaction is treated as a tool invocation, and the corresponding external feedback is represented as a tool observation. Under this abstraction, heterogeneous interfaces share a single invocation path in the rollout loop, and the framework generalizes to new interaction types without changes to the rollout logic.

\paragraph{Fully Asynchronous Rollout}
Agent rollout interleaves token generation with tool calling, and the time cost of each step varies greatly across turns and tasks. Batch-synchronous designs block generation to collect requests and launch tool executions, creating bubbles during rollout (\Cref{fig:async}, right). We instead adopt a fully asynchronous design for the entire rollout: as shown in \Cref{fig:async} (left), token generation, tool calling, and reward calculation each proceed asynchronously. The resource system is likewise asynchronous, so resource allocation and scaling overlap with ongoing rollouts rather than blocking them.

\subsection{System Overview}\label{sec:system}
We introduce the structure of our framework in this section, which consists of four layers to make agentic RL easy to use while maintaining efficiency. The context layer and resource layer combine to form a unified resource system.

\subsubsection{Agent Layer}\label{sec:agent_layer}
The agent layer exposes the main interfaces to users and consists of three modules: the agent, tools, and rewards. Together they decompose agentic RL into three definitions, while rollout execution and resource management are handled by the lower layers.

\paragraph{Agent}
An agent is the entity that completes tasks: it generates actions, invokes tools, and may perform additional operations such as querying its memory. The framework provides common functionality: running the rollout loop, parsing tool calls, and collecting trajectories. A working agent requires only a model and a set of tools. Beyond the defaults, users can override these behaviors or extend agents with custom components, such as a memory module.

\begin{figure}[t]
\captionsetup{type=lstlisting}
\begin{minipage}[t]{0.49\linewidth}
\begin{codebox}
\begin{Verbatim}[commandchars=\\\{\},fontsize=\footnotesize]
\ckw{@tool}(name=\cst{"run_shell_command"})
\ckw{async def} \cfn{run_shell_command}(
        cmd: str, context: Context) -> str:
    \ccm{# declare the resource it needs}
    \hlres{spec = ContainerResourceSpec(}
        \hlres{category=\cst{"container"},}
        \hlres{image=\cst{"ubuntu:22.04"},}
        \hlres{max_global_num=24)}
    \ccm{# acquire it through the context layer}
    \hlctx{box = \ckw{await} context.\cfn{acquire_resource}(}
        \hlctx{id=context.rollout_id,}
        \hlctx{spec=spec, scope=\cst{"rollout"})}
    \ckw{return await} box.\cfn{run_cmd}(cmd, timeout=60)
\end{Verbatim}
\end{codebox}
\end{minipage}\hfill
\begin{minipage}[t]{0.49\linewidth}
\begin{codebox}
\begin{Verbatim}[commandchars=\\\{\},fontsize=\footnotesize]
\ckw{@reward}(name=\cst{"llm_judge_reward"})
\ckw{async def} \cfn{llm_judge_reward}(
        final_response: str, answer: str,
        context: Context) -> float:
    \ccm{# declare the resource it needs}
    \hlres{spec = APIModelResourceSpec(}
        \hlres{model_name_or_path=\cst{"Qwen3-32B"},}
        \hlres{host=\cst{"judge-server"}, port=8000)}
    \ccm{# acquire it through the context layer}
    \hlctx{judge = \ckw{await} context.\cfn{acquire_resource}(}
        \hlctx{id=\cst{"judge"}, spec=spec, scope=\cst{"global"})}
    out = \ckw{await} judge.\cfn{generate_async}(
        [judge_prompt(final_response, answer)])
    \ckw{return} float(\cst{"yes"} \ckw{in} out[0].lower())
\end{Verbatim}
\end{codebox}
\end{minipage}
\caption{One acquisition interface for two resource kinds. \protect\hlres{Orange lines} declare the resource a function needs; \protect\hlctx{green lines} acquire it through the context layer. \emph{Left:} a tool that runs shell commands in an isolated container whose pool holds at most 24 instances; acquiring under the rollout id makes all turns of a rollout reuse one container, released when the rollout ends. \emph{Right:} a reward that judges the final answer with an external model API; the same call acquires a model resource with global scope, shared across rollouts (\texttt{judge\_prompt} builds the chat messages). Neither function contains allocation or cleanup logic.}
\label{lst:tool}
\end{figure}

\paragraph{Tool \& Reward}
A tool is defined by decorating a function or inheriting from the base class. The challenge is that agent tasks vary widely in their execution requirements: a calculator tool is a stateless local function, whereas an agent exploring an operating system needs its commands executed in an isolated environment, concurrently with many other rollouts. To cover both cases under one definition style, a tool declares the resources it needs and obtains them through the context layer, which communicates with the resource layer on its behalf; the tool body itself remains a plain Python function. \Cref{lst:tool} shows a tool that acquires a container from the context and runs a command inside it. Rewards generally follow the same design as tools: a reward function is a decorated function that can take information or acquire resources from the context layer.

\subsubsection{Rollout Layer}\label{sec:rollout_layer}
The rollout layer composes the agent-layer components into the agent loop: it drives the alternation between LLM generation and tool execution, invokes the reward function on the finished trajectory, and transfers rollout-level information to the context layer.
\name currently provides two rollout strategies to support multi-turn and multi-segment training. \emph{Chain rollout} keeps one growing conversation per trajectory, where the model is trained on all of its assistant turns in a single sequence. \emph{Step rollout} re-prompts the model at every environment step with a fresh prompt that renders the task and a window of the most recent observations and actions, and produces one training row per step; this bounds the context length and gives each step its own sample, which step-level credit assignment such as GiGPO~\citep{feng2025group} requires. Other rollout patterns, such as tree search, can be added by registering a new loop. At the start of a rollout, the layer registers rollout information, such as input data and rollout identifiers, with the context layer, and at the end, it signals completion so that rollout-scoped resources can be reclaimed.

\begin{figure}[t]
\centering
\begin{minipage}[t]{0.48\linewidth}
\centering
\includegraphics[width=\linewidth]{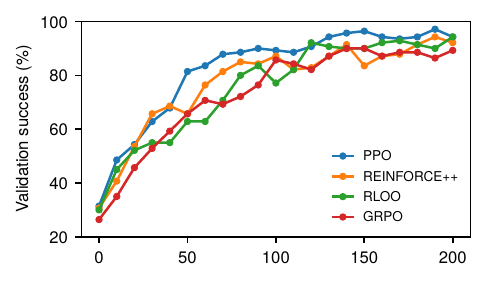}
\caption{ALFWorld validation success of Qwen2.5-3B-Instruct trained with PPO, REINFORCE++, RLOO, and GRPO under chain rollout.}
\label{fig:rl_algorithms}
\end{minipage}\hfill
\begin{minipage}[t]{0.48\linewidth}
\centering
\includegraphics[width=\linewidth]{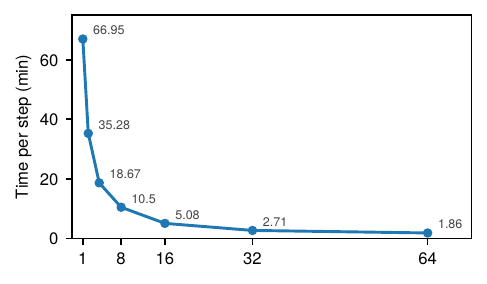}
\caption{ScienceWorld training time per step as the number of containers increases from 1 to 64.}
\label{fig:resource_scaling}
\end{minipage}
\end{figure}

\subsubsection{Context Layer}\label{sec:context_layer}
The context layer sits between the rollout and resource layer, and serves two roles. First, it \emph{injects information} from inputs (e.g., reference answers, rollout identifiers) into tools and rewards. Second, it \emph{brokers resources}: when a tool or reward requests a resource, the context layer forwards the request to the resource layer, hands the acquired instance to the caller, and tracks the association between resources and rollouts. This enables lifecycle management: when a rollout finishes, the context layer releases the resources acquired, while shared resources remain available for other rollouts. Tool and reward definitions thus stay free of allocation, communication, and cleanup logic, which is what keeps the agent-layer interfaces in \Cref{lst:tool} plain functions.

\subsubsection{Resource Layer}\label{sec:environment_management}
The resource layer performs allocation and recycling of the resources that tools and rewards depend on, and is designed to let environments scale with training throughput. We define a \emph{resource} to be the smallest unit used by a tool or reward, and represent each kind of resource as a distinct, typed specification resolved through a single engine: containers (image, CPU and memory limits, isolation backend), local model services (model, parallelism, and memory configuration for a vLLM server), external model APIs (endpoint and credentials), and in-process environments (e.g., a Chess engine). All kinds implement a common lifecycle interface (\texttt{start}, \texttt{reset}, \texttt{status}, \texttt{release}), so one scheduler manages heterogeneous resources rather than a single kind.

The engine maintains a pool per resource specification. Each resource is acquired under a caller-supplied identifier, and a repeated acquisition with the same identifier returns the same live instance instead of allocating a new one. A container created in the first turn of a rollout is therefore reused across all subsequent turns, amortizing setup cost over the trajectory. Each pool declares a maximum capacity; when the pool is saturated, further acquisitions wait asynchronously until other rollouts release instances, which bounds total resource usage under high rollout concurrency without blocking generation. Resources are scoped either to a single rollout that is released automatically when the rollout ends, or globally, persisting across rollouts to serve as shared services.

Resource specifications are decoupled from where resources physically run: \emph{runners} launch the same specification on different substrates, and we currently provide Local, Slurm, and Ray launchers, which run on local machines, Slurm jobs, and Ray clusters, respectively. The container isolation backend is likewise pluggable, so lighter-weight isolation mechanisms can replace full containers without changing the scheduling logic. We use enroot as our default container backend. To make it fully compatible with our design, we also write a Python-based asynchronous client for it\footnote{\url{https://github.com/Agent-One-Lab/enroot-py}}.

\subsection{Pre-Built Tools and Tasks}\label{sec:prebuilt_tools_tasks}
To facilitate the use of our framework, we have built tools and integrated agent tasks. Tasks include SearchQA \citep{jin2025search}, WebShop \citep{yao2022webshop}, ALFWorld \citep{shridhar2021alfworld}, ScienceWorld \citep{wang2022scienceworld}, and SimuScene \citep{wang2026simuscene}. We also provide OS-related tools, including file reading, writing, editing, file search, Python, and general bash. We use these tools to support SWE-Bench \citep{jimenez2024swe} style RL training.

\section{Experiments} \label{sec:experiments}
We detail our experiments in this section, where we first train with different RL algorithms for multi-turn agents, then present experimental results over different tools and environments.

\subsection{Agentic RL with Different Algorithms}

To evaluate how different reinforcement learning algorithms behave on \name, we train Qwen2.5-3B-Instruct~\citep{qwen2025qwen25technicalreport} on ALFWorld~\citep{shridhar2021alfworld} with four algorithms: PPO~\citep{schulman2017proximal}, REINFORCE++~\citep{hu2501reinforce++}, RLOO~\citep{ahmadian-etal-2024-back}, and GRPO~\citep{shao2024deepseekmath}. All runs use chain rollout and train on the sampled token IDs, with otherwise identical settings: 16 tasks with 8 trajectories each per step, up to 50 turns per episode, a learning rate of $5\times 10^{-7}$, and 200 training steps. PPO additionally trains a critic and estimates advantages with GAE; since GAE and REINFORCE++ discount per token, both use a discount factor of 1. We evaluate success on the \texttt{valid\_seen} split every 10 steps with sampling temperature 0.4. All runs are conducted on a single node with 8 H200 GPUs.

As shown in \Cref{fig:rl_algorithms} and \Cref{tab:alfworld_algos} (\Cref{app:detailed_results}), all four algorithms raise the success rate from 26--31\% to 88--95\%. PPO learns fastest, first reaching 80\% success at step 50 compared with steps 70, 80, and 100 for REINFORCE++, RLOO, and GRPO, and ends highest at 95.2\%; GRPO ends lowest at 88.1\%, although it also starts lowest. This is consistent with its learned value function giving denser credit assignment over long multi-turn trajectories, at the cost of training a critic. All four runs take 12--16 hours, since rollout dominates the step time and steps become faster as the agent learns to finish tasks in fewer turns.

\subsection{Agentic RL for Tools with Resources}

\begin{figure*}[t]
\centering
\includegraphics[width=\linewidth]{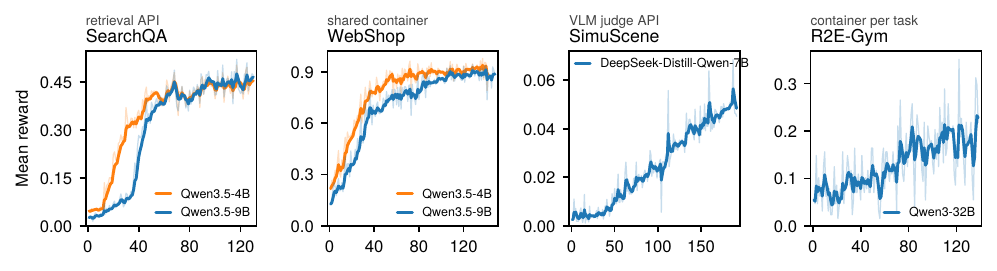}
\caption{Training reward on four tasks that use different kinds of resources: a retrieval API (SearchQA), a container shared across rollouts (WebShop), a VLM judge API (SimuScene), and one container per task (R2E-Gym). Bold lines are exponential moving averages; faint lines are per-step values.}
\label{fig:experiments_resource}
\end{figure*}

\paragraph{Experimental Setup}
We next conduct experiments on the tools and environments introduced in \Cref{sec:prebuilt_tools_tasks}, on H200 nodes with 8 GPUs each. We use GRPO \citep{shao2024deepseekmath} as the default algorithm for each task. For SearchQA, retrieval APIs are deployed and used as resources. For WebShop, we wrap it into a container shared across rollouts. We use Qwen3.5 \citep{team2026qwen3} for these two tasks. For SimuScene \citep{wang2026simuscene}, resources are Vision-Language Model APIs, which are used to judge answer correctness. We use DeepSeek-Distill-Qwen-7B \citep{guo2025deepseek} for training. SWE-Bench \citep{jimenez2024swe} is the heaviest setting: we use the R2E-Gym \citep{jain2025r2e} dataset for training. Each task uses a single container as the resource, executes commands, and stops it after the reward is calculated. We use Qwen3-32B for training, with at most 30 turns and a context of about 40k tokens. The training settings are listed in \Cref{tab:resource_tasks_setup} and the reward designs in \Cref{app:tool}.

As shown in \Cref{fig:experiments_resource}, the rewards for SearchQA, WebShop, and R2E-Gym first increase and then plateau as training progresses. The reward for SimuScene is still increasing after 160 steps. These results demonstrate that \name is able to train the newest Qwen3.5 series of models, reasoning model (DeepSeek-Distill-Qwen-7B) and large model (Qwen3-32B) with a 40k-token context and up to 30 turns. We also present experiments on the Qwen2.5 series across more tasks in \Cref{app:more_experiments}.

\section{Analysis}

\subsection{Scaling Resources}

\begin{figure}[t]
\centering
\includegraphics[width=\linewidth]{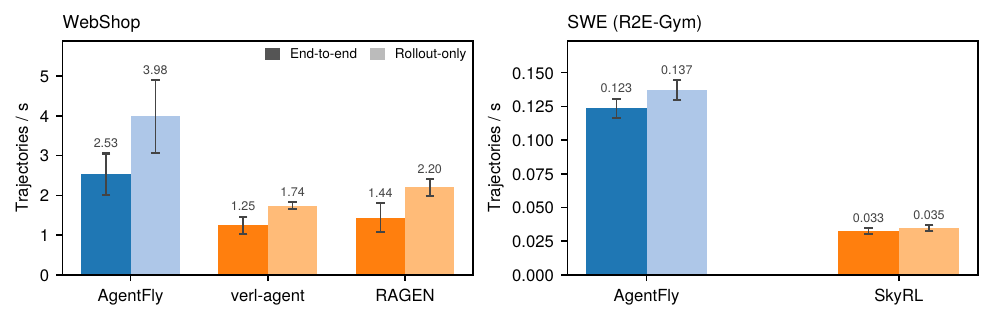}
\caption{Training throughput (trajectories/s) on WebShop and R2E-Gym, with all frameworks on 8 GPUs. End-to-end includes the policy update; rollout-only covers generation and environment interaction. Error bars show one standard deviation across steps.}
\label{fig:throughput}
\end{figure}

We then scale the resources to show the effectiveness of our resource system. As shown in \Cref{lst:tool}, we can configure the number of containers in the tool definition, and the context layer will acquire them from the low-level resource layer. We use the ScienceWorld task as an example. Except for the resource number, all settings are fixed (\Cref{app:detailed_results}). As shown in \Cref{fig:resource_scaling}, when the number of resources increases, the time needed for training decreases and then remains stable. When enough resources are provided, the training becomes mainly bottlenecked by the generation throughput and RL trainer's parameter updating.

\subsection{Throughput Comparison}\label{sec:throughput}

We provide the first cross-framework throughput comparison between different agentic RL frameworks. We select WebShop \citep{yao2022webshop} and R2E-Gym \citep{jain2025r2e}, both of which are officially supported by \name and the frameworks we compare. Since we are using enroot while SkyRL's \citep{cao2025skyrl} default container backend is Docker, we also implement an enroot backend for SkyRL for a fair comparison. We report trajectory throughput in two settings: end-to-end training, which measures the throughput of the whole system; and rollout-only, which excludes the RL trainer's effects and focuses on the agent module's rollout only. For WebShop, we train Qwen2.5-3B-Instruct on each framework for 150--200 steps (\Cref{tab:webshop_throughput_setup}). For R2E-Gym, we train Qwen3-4B with a context of about 30k tokens and at most 20 turns, and measure the first ten training steps (\Cref{tab:swe_setup}). Other hyperparameters (e.g. batch size, group size) are kept the same. Throughput is calculated based on the average value of one step.

As shown in \Cref{fig:throughput}, in both WebShop and R2E-Gym, \name achieves the best throughput. More specifically, it achieves around 2 times the performance of verl-agent, 1.75 times that of RAGEN \citep{ragen}. The advantage is larger on R2E-Gym, where \name achieves 3.8 times the throughput of SkyRL. This demonstrates the efficiency of our dedicated resource system.

\subsection{Token Drift in Multi-Turn Training}\label{sec:token_drift}

\begin{figure}[t]
\centering
\includegraphics[width=\linewidth]{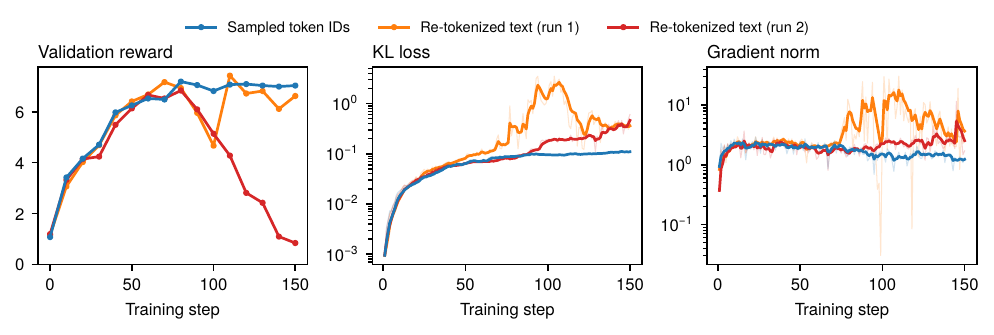}
\caption{Training on sampled token IDs versus re-tokenized text on WebShop with Qwen3.5-4B (chain rollout, GRPO). The two runs differ only in whether the sampled token IDs are kept; the re-tokenized setting is shown for two runs. KL loss and gradient norm are exponential moving averages over per-step values (faint), on a log scale.}
\label{fig:token_drift}
\end{figure}

A multi-turn agent decodes each response to text so that tool calls can be parsed and executed, and a naive pipeline later re-tokenizes that text for training. However, re-tokenization is not an identity transformation for BPE tokenizers, so the trainer computes log-probabilities, importance ratios, and the KL penalty on tokens the policy never produced, causing drift between inference and training. In multi-turn rollouts the drift also enters later prompts, since earlier responses become part of the context. Such drift is harmful to training, as shown in \Cref{fig:token_drift}.
\name keeps token IDs as the unit of exchange between inference and training. Before generation, the response IDs from previous turns are directly used to form prompt IDs. Generated response IDs are saved for later generation and training. During training, the saved token IDs are directly used to form the trajectory, so there is no token drift between generation and training. This is supported through a separate library for LLM chat template\footnote{\url{https://github.com/Agent-One-Lab/chat-bricks}}.

To measure the effect, we train Qwen3.5-4B on WebShop with chain rollout and GRPO for 150 steps, one run with sampled token IDs and two runs with every turn re-encoded, keeping all other settings identical. We observe that token drift affects 36\% of trajectories on average, but only 0.07\% of tokens. This small drift ratio is nonetheless enough to destabilize training (\Cref{fig:token_drift}). All three runs track each other for about 75 steps. The re-tokenized runs then diverge. The KL loss peaks at 3.5 and 0.6 and gradient norms at 30 and 12, compared with at most 0.12 and 2.9 when training on IDs, and policy entropy drops sharply. One re-tokenized run recovers only partially and ends at a validation reward of 6.6, and the other collapses from 6.8 to 0.8, while training on sampled IDs stays stable and ends at 7.0.

\subsection{Chain versus Step Rollout}\label{sec:chain_step}

Because both rollout strategies run on the same agent, tools, and resources, \name can compare them under otherwise identical settings. We train Qwen2.5-Instruct models of three sizes on WebShop for 150 steps and a 1.5B model on ALFWorld for 200 steps, with GRPO and with GiGPO~\citep{feng2025group}, changing only the rollout strategy (\Cref{tab:chain_step} in \Cref{app:detailed_results}).

Chain rollout matches or outperforms step rollout in nearly every setting. Its advantage is largest for GRPO, where in-distribution WebShop success rises by 5.8, 19.1, and 2.2 points at 1.5B, 3B, and 7B, and it carries over to held-out products at 3B and 7B (52.5 vs.\ 39.8 and 58.0 vs.\ 48.0). GiGPO's step-level credit assignment, in contrast, helps most under step rollout, where it improves on GRPO by 6.4, 15.9, and 7.6 points in distribution; under chain rollout the gain shrinks to 1.6, 0.9, and 5.7 points. The full interaction history therefore substitutes for much of what per-step grouping provides. ALFWorld shows the same pattern: chain rollout improves GRPO from 75.5\% to 91.0\% and GiGPO from 89.3\% to 98.1\%, and GiGPO's gain over GRPO shrinks from 13.8 points under step rollout to 7.1 points under chain rollout. At 1.5B the held-out WebShop results of all four configurations lie within about one point of each other, within the evaluation noise of roughly two points. The best rollout strategy thus depends on the task, the model size, and the estimator, which is why \name supports both behind the same interface.

\section{Related Work}

\begin{table}[t]
\centering
\begin{tblr}{
width=\linewidth,
colspec={lcccll},
row{1}={font=\bfseries},
hlines, hline{1,Z}={1pt}, hline{2}={0.6pt}, row{2}={bg=black!6},
}
Framework        & Agent Abs.\textsuperscript{a} & Async & Resource Scaling\textsuperscript{b} & Multi-Modal\textsuperscript{c} & Backends \\
\textbf{AgentFly} & \fullcirc  & \fullcirc  & \fullcirc  & T+I              & verl \\
RAGEN             & \emptycirc & \emptycirc & \halfcirc  & \emptycirc       & verl \\
verl-agent        & \emptycirc & \emptycirc & \halfcirc  & T+I              & verl \\
RL-Factory        & \halfcirc  & \fullcirc  & \emptycirc & T+I              & verl \\
agent-lightning   & \halfcirc  & \fullcirc  & \emptycirc & N/A              & multiple\textsuperscript{d} \\
verl-tool         & \halfcirc  & \fullcirc  & \emptycirc & T+I+V            & verl \\
SkyRL-Agent       & \fullcirc  & \fullcirc  & \halfcirc  & \emptycirc       & multiple\textsuperscript{e} \\
miles             & \emptycirc & \fullcirc  & \halfcirc  & T+I              & slime\textsuperscript{f} \\
\end{tblr}
\caption{
Comparison of agentic RL frameworks.
\fullcirc{} full / native, \halfcirc{} partial or depends on configuration, \emptycirc{} none.
\textsuperscript{a}~Agent abstraction including agent, tool, and reward abstractions.
\textsuperscript{b}~\halfcirc{} indicates inherited from underlying infrastructure (e.g., Ray, SkyRL-Agent on K8s);
\fullcirc{} indicates dedicated resource system (e.g., AgentFly's resource engine).
\textsuperscript{c}~T = text, I = image, V = video.
\textsuperscript{d}~LangChain, AutoGen, CrewAI, OpenAI SDK for agent rollout.
\textsuperscript{e}~SkyRL-train, verl, Tinker. \textsuperscript{f}~slime (SGLang for rollout, Megatron-LM for training).
}
\label{tab:comparison}
\end{table}

There are several emerging RL training frameworks, some of which are mainly designed for LLMs and have agentic RL as an extension. Among them, \texttt{verl} \citep{sheng2024hybridflow} uses hybrid engines to train LLMs efficiently. AReaL \citep{fu2025areal} orchestrates rollout and updating to be fully asynchronous with better efficiency, while suffering from the off-policy problem. slime \citep{slime_github} focuses on large-scale RL training, and has been validated on GLM series post-training \citep{glm52_blog}. Building on slime, miles \citep{miles_github} further pushes large-scale training through a unified FP8 rollout--training pipeline.
However, such frameworks focus on training design and lack a systematic architecture for agentic RL, making direct comparisons difficult.

We mainly compare with frameworks specifically designed for agentic RL.
RAGEN \citep{ragen} proposes StarPO in the form of a State-Thinking-Actions-Reward pipeline, but offers limited support for tool and environment extension and scaling. RL-Factory \citep{rlfactory} features easy and efficient training, which integrates MCP-based tool definitions and asynchronous parallel tool calls. verl-agent \citep{feng2025group} converts multi-turn agent trajectories into single-turn episodes through context appending, thereby converting multi-turn RL into multiple single-turn RL. They use Ray to launch multi-process environments, while making the rollout synchronous. Agent-Lightning \citep{luo2025agent} supports training with multiple agent front-ends, so migration cost is minimal. However, these front-ends are not designed for agentic RL, and suffer from isolation, efficiency, and scaling problems. SkyRL-Agent \citep{cao2025skyrl} targets long-horizon agents such as software-engineering agents, with asynchronous dispatching of rollouts, lightweight tool integration, and several training backends (SkyRL-train, \texttt{verl}, and Tinker). Its environments run in per-task sandbox runtimes managed outside the trainer, whereas \name schedules containers, model services, and APIs through one resource engine. Since different frameworks adopt different designs, we present a more systematic comparison in \Cref{tab:comparison}, where we compare (I) abstraction level, (II) asynchronous support, (III) environment scaling, (IV) multi-modality support, and (V) backends support.

Overall, \name balances ease of development with efficiency.

\section{Conclusion}
We presented \name, an agentic RL framework built around a unified resource system. Containers, local model services, external model APIs, and in-process environments are scheduled as typed resources through one engine, with lifecycle management and asynchronous running supported. On top of this system, a four-layer design lets users define agents, tools, and rewards as plain functions, while rollout composition, context management, and resource handling stay in the lower layers. Our experiments show what this design enables. \name trains models from 1.5B to 32B parameters on tasks that rely on different kinds of resources, from retrieval and judge APIs to shared and per-task containers, and supports multiple RL algorithms. Scaling the container pool reduces training time per step, and at equal hardware \name delivers about 2-4$\times$ the training throughput of other agentic RL frameworks. Training on sampled token IDs avoids the drift that destabilizes multi-turn training when responses are re-tokenized, and supporting both chain and step rollout behind one interface shows that the better strategy depends on the task, model size, and estimator. We hope \name provides a practical foundation for scaling agentic RL to richer environments.

\bibliography{anthology-1,anthology-2,custom}
\bibliographystyle{plainnat}

\appendix
\newpage
\clearpage

\section{Multi-Turn RL Training}\label{app:multiturn}
Formally, we define an agent trajectory with $k$ turns as $T_k=(p, (r_1, o_1), (r_2, o_2), \ldots, (r_k))$, where $p$ is the system and user prompt, $r_i$ denotes the LLM-generated response, and $o_i$ denotes the observation from tool outputs. For a single-turn trajectory $T_1=(p, (r_1))$, which is the standard case of RL for LMs, the PPO loss can be written as a function of the current token $a_t$, its context $s_t$, and the advantage $\hat{A}_t$:
\begin{equation*}
\mathcal{L}_{PPO}(\theta) = \sum^{L}_{t=1}f(a_t, s_t, \hat{A}_t) \cdot M_t
\end{equation*}
where $M_t = 1$ if $a_t \in r_1$ and $M_t=0$ otherwise. To extend training to multi-turn trajectories, we generalize the mask to cover all LLM-generated turns:
\begin{equation*}
M_t = \left\{
\begin{array}{ll}
1 & \text{if } a_t \in \{r_1, r_2, \ldots, r_k\} \\
0 & \text{otherwise}
\end{array}
\right.
\end{equation*}
In our implementation, we also apply the mask to the advantage $\hat{A}$ to avoid unnecessary computation.

To make chat template rendering and multi-turn mask construction easy and efficient, we built a separate library\footnote{\url{https://github.com/Agent-One-Lab/chat-bricks}} and integrated it into \name.

\section{Detailed Setup and Results}\label{app:detailed_results}
This section gives the training settings of the experiments in \Cref{sec:experiments} and the analysis in \Cref{sec:throughput,sec:token_drift,sec:chain_step}, together with their full results. All runs use \texttt{verl} with FSDP for training and vLLM for generation. Unless stated otherwise, runs use a single node with 8 H200 GPUs, the policy loss uses a clip ratio of 0.2 and token-mean aggregation, rollouts are sampled at temperature 1.0, and the model is trained on the sampled token IDs.

\paragraph{RL algorithms on ALFWorld} The four runs in \Cref{sec:experiments} train Qwen2.5-3B-Instruct with the ALFWorld chain-rollout settings of \Cref{tab:webshop_alfworld_setup} (16 tasks $\times$ 8 trajectories per step, at most 50 turns of 512 tokens, a 16{,}384-token context, learning rate $5\times10^{-7}$, a mini-batch of 8 trajectories, low-variance KL loss with coefficient 0.01, entropy coefficient 0.001, and 200 steps), on one node with 8 H200 GPUs. Only the advantage estimator changes. PPO trains a critic initialized from the policy model with learning rate $10^{-5}$ and estimates advantages with GAE ($\lambda=1$, $\gamma=1$); REINFORCE++ uses $\gamma=1$; RLOO uses the leave-one-out baseline within each group of 8 trajectories; GRPO uses \name's episode-level estimator with a per-step discount of 0.95 and an invalid-action penalty of 0.1. The reward is 10 for a successful episode and 0 otherwise. \Cref{tab:alfworld_algos} summarizes the results.

\begin{table}[t]
\centering
\caption{ALFWorld results of Qwen2.5-3B-Instruct with different RL algorithms. Success is measured on \texttt{valid\_seen}; Final is the mean of the last three evaluations (steps 180--200), and Steps to 80\% is the first evaluation at or above 80\% success. Time is the training time of the 200-step run on 8 H200 GPUs.}
\label{tab:alfworld_algos}
\begin{tabular}{lccccc}
\toprule
Algorithm & Initial (\%) & Final (\%) & Best (\%) & Steps to 80\% & Time (h) \\
\midrule
PPO          & 31.4 & \textbf{95.2} & \textbf{97.1} & \textbf{50} & 12.5 \\
REINFORCE++  & 30.7 & 92.6 & 94.3 & 70 & 12.9 \\
RLOO         & 30.0 & 91.9 & 94.3 & 80 & 14.2 \\
GRPO         & 26.4 & 88.1 & 90.0 & 100 & 16.2 \\
\bottomrule
\end{tabular}
\end{table}

\begin{table}[t]
\centering
\caption{Success rate (\%) of chain and step rollout with GRPO and GiGPO. WebShop: in-distribution success on the 512 validation tasks of \citet{feng2025group} (mean of the last three evaluations) and success on 512 held-out tasks whose products never appear in training (final checkpoint). ALFWorld: success on \texttt{valid\_seen} (mean of the last three evaluations). All models are Qwen2.5-Instruct.}
\label{tab:chain_step}
\begin{tabular}{llccccccc}
\toprule
 & & \multicolumn{3}{c}{WebShop (in-distribution)} & \multicolumn{3}{c}{WebShop (held-out)} & ALFWorld \\
\cmidrule(lr){3-5}\cmidrule(lr){6-8}\cmidrule(lr){9-9}
Rollout & Algorithm & 1.5B & 3B & 7B & 1.5B & 3B & 7B & 1.5B \\
\midrule
Step  & GRPO  & 58.9 & 50.1 & 65.8 & \textbf{53.1} & 39.8 & 48.0 & 75.5 \\
Step  & GiGPO & 65.3 & 66.0 & 73.4 & 52.1 & 46.7 & 59.0 & 89.3 \\
Chain & GRPO  & 64.7 & 69.2 & 68.0 & 52.3 & 52.5 & 58.0 & 91.0 \\
Chain & GiGPO & \textbf{66.3} & \textbf{70.1} & \textbf{73.7} & 52.5 & \textbf{55.3} & \textbf{61.1} & \textbf{98.1} \\
\bottomrule
\end{tabular}
\end{table}

\paragraph{Tasks with different resources} \Cref{tab:resource_tasks_setup} lists the settings of the four tasks in \Cref{fig:experiments_resource}. All runs use GRPO and the multi-turn agent loop; SimuScene is single-turn, since the policy writes one simulation program that is rendered to a video and judged by vision-language models.

\begin{table*}[t]
\centering
\small
\caption{Settings of the tasks with different resources (\Cref{fig:experiments_resource}). Tasks $\times$ group size is the number of trajectories per step.}
\label{tab:resource_tasks_setup}
\resizebox{\linewidth}{!}{\begin{tabular}{lcccc}
\toprule
 & SearchQA & WebShop & SimuScene & R2E-Gym \\
\midrule
Model & Qwen3.5-\{4B, 9B\}-Base & Qwen3.5-\{4B, 9B\} & DeepSeek-R1-Distill-Qwen-7B (SFT) & Qwen3-32B \\
Resource & retrieval API & shared container & VLM judge API & container per task \\
Training data & NQ + HotpotQA & WebShop (small) & SimuScene & R2E-Gym-Lite \\
Reward & exact match & WebShop score & VLM-judged pass & unit tests \\
Tasks $\times$ group size & $256\times 5$ & $32\times 8$ & $32\times 8$ & $16\times 8$ \\
Max turns & 3 & 15 & 1 & 30 \\
Max tokens per turn & 512 & 384 & 14{,}336 & 4{,}096 \\
Max context length & 16{,}384 & 16{,}384 & 16{,}384 & 40{,}960 \\
Learning rate & $5\times 10^{-7}$ & $5\times 10^{-7}$ & $5\times 10^{-7}$ & $4\times 10^{-7}$ \\
Mini-batch (trajectories) & 1{,}280 & 256 & 128 & 128 \\
KL loss coefficient & 0.002 (MSE) & 0.001 (MSE) & -- & 0.001 (MSE) \\
Entropy coefficient & 0.002 & 0.001 & 0.001 & 0.001 \\
GPUs (H200) & 8 & 16 & 16 & 32 \\
Steps shown & 130 & 142--148 & 191 & 138 \\
\bottomrule
\end{tabular}}
\end{table*}

\paragraph{Resource scaling on ScienceWorld} The runs in \Cref{fig:resource_scaling} train Qwen2.5-3B-Instruct on ScienceWorld with GRPO on one node with 8 H200 GPUs: 64 tasks $\times$ 8 trajectories per step, at most 20 turns of 256 tokens, a 16{,}384-token context, learning rate $4\times10^{-7}$, a mini-batch of 512 trajectories, KL loss coefficient 0.001 (MSE), and entropy coefficient 0.001. Each rollout acquires its own ScienceWorld container, and the only setting that changes between runs is the maximum number of containers in the pool, from 1 to 64.

\paragraph{Throughput on WebShop} \Cref{tab:webshop_throughput_setup} lists the settings of the WebShop throughput comparison (\Cref{sec:throughput}). All three frameworks train Qwen2.5-3B-Instruct with GRPO on 8 H200 GPUs with 32 tasks $\times$ 8 trajectories per step and the same learning rate, KL, and entropy coefficients; each uses its own WebShop environment and agent loop. Throughput is the number of trajectories per step divided by the step time (end-to-end) or by the rollout time (rollout-only), averaged over all training steps after five warm-up steps.

\begin{table}[t]
\centering
\small
\caption{Settings of the WebShop throughput comparison.}
\label{tab:webshop_throughput_setup}
\begin{tabular}{lccc}
\toprule
 & \name & verl-agent & RAGEN \\
\midrule
Model & \multicolumn{3}{c}{Qwen2.5-3B-Instruct} \\
Algorithm & \multicolumn{3}{c}{GRPO} \\
Tasks $\times$ group size & \multicolumn{3}{c}{$32\times 8$} \\
Max turns & 15 & 15 & 15 (9 actions) \\
Max tokens per turn & 384 & 384 & 400 \\
History in context & full & last 2 turns & full \\
Max context length & 16{,}384 & 8{,}192 (prompt) & 15{,}000 \\
Learning rate & \multicolumn{3}{c}{$5\times 10^{-7}$} \\
Mini-batch & 256 & 64 (step samples) & 256 \\
KL loss coefficient & 0.001 (MSE) & 0.001 (low-var.) & 0.001 (low-var.) \\
Entropy coefficient & \multicolumn{3}{c}{0.001} \\
Rollout tensor parallelism & 1 & 2 & 1 \\
GPUs (H200) & \multicolumn{3}{c}{8} \\
\bottomrule
\end{tabular}
\end{table}

\paragraph{Throughput on R2E-Gym} \Cref{tab:swe_setup} lists the settings of the R2E-Gym throughput comparison (\Cref{sec:throughput}). Both frameworks train Qwen3-4B with GRPO on R2E-Gym-Lite with 64 trajectories per step, at most 20 turns, 4{,}096 generated tokens per turn, and a context of about 30k tokens, and run each task in an enroot container; for SkyRL we use our enroot backend in place of its default Docker backend. We use each framework's own agent scaffold: \name gives the agent file-editing, search, and Python-execution tools, while SkyRL uses its mini-SWE-agent with a bash tool. Throughput is the number of trajectories per step divided by the step time (end-to-end) or by the generation time (rollout-only), averaged over the training steps after two warm-up steps.

\begin{table}[t]
\centering
\small
\caption{Settings of the R2E-Gym throughput comparison.}
\label{tab:swe_setup}
\begin{tabular}{lcc}
\toprule
 & \name & SkyRL \\
\midrule
Model & \multicolumn{2}{c}{Qwen3-4B} \\
Algorithm & \multicolumn{2}{c}{GRPO} \\
Tasks $\times$ group size & $8\times 8$ & $16\times 4$ \\
Trajectories per step & \multicolumn{2}{c}{64} \\
Max turns & \multicolumn{2}{c}{20} \\
Max tokens per turn & \multicolumn{2}{c}{4{,}096} \\
Max context length & 30{,}000 & 30{,}720 \\
Learning rate & $4\times 10^{-7}$ & $1\times 10^{-6}$ \\
Mini-batch (trajectories) & \multicolumn{2}{c}{64} \\
KL loss coefficient & 0.001 (MSE) & 0.001 (k3) \\
Entropy coefficient & 0.001 & 0 \\
Inference engine & \multicolumn{2}{c}{vLLM, 4 replicas $\times$ TP 2} \\
Container runtime & \multicolumn{2}{c}{enroot} \\
Agent tools & file edit, search, Python & bash \\
\bottomrule
\end{tabular}
\end{table}

\paragraph{Token drift and chain versus step rollout} \Cref{tab:webshop_alfworld_setup} lists the settings of the token-drift runs (\Cref{sec:token_drift}) and the chain and step rollout runs (\Cref{sec:chain_step}). WebShop uses the 2{,}400 training and 512 validation goals replayed from \citet{feng2025group}; ALFWorld uses 3{,}553 training games and the 140 \texttt{valid\_seen} games. GRPO and GiGPO share one implementation: GRPO sets the step-level advantage weight to 0 and normalizes episode advantages by the group mean and standard deviation, while GiGPO adds step-level advantages with weight 1 and discount 0.95. The step-rollout runs follow the recipe of \citet{feng2025group}: each step sees the task and the last two turns, and the learning rate and mini-batch are set per step sample. The chain-rollout runs keep the full interaction history in one sequence and use a lower learning rate and a mini-batch of whole trajectories. Qwen3.5-4B runs with its default thinking mode enabled, so it is given more tokens per turn. Validation samples one trajectory per task at temperature 0.4 every 10 steps; the held-out WebShop set of 512 goals is evaluated once at the final checkpoint with the same settings. The two token-drift settings differ only in whether the sampled token IDs are kept.

\begin{table*}[t]
\centering
\small
\caption{Settings of the token-drift runs (\Cref{sec:token_drift}) and the chain and step rollout runs (\Cref{sec:chain_step}). Mini-batches count whole trajectories for chain rollout and single-step samples for step rollout.}
\label{tab:webshop_alfworld_setup}
\begin{tabular}{lccccc}
\toprule
 & Token drift & \multicolumn{2}{c}{WebShop} & \multicolumn{2}{c}{ALFWorld} \\
\cmidrule(lr){2-2}\cmidrule(lr){3-4}\cmidrule(lr){5-6}
 & WebShop & Step & Chain & Step & Chain \\
\midrule
Model & Qwen3.5-4B & \multicolumn{2}{c}{Qwen2.5-\{1.5B, 3B, 7B\}-Instruct} & \multicolumn{2}{c}{Qwen2.5-1.5B-Instruct} \\
Algorithm & GRPO & \multicolumn{2}{c}{GRPO, GiGPO} & \multicolumn{2}{c}{GRPO, GiGPO} \\
GiGPO normalization & -- & \multicolumn{2}{c}{mean} & \multicolumn{2}{c}{mean and std} \\
Tasks $\times$ group size & \multicolumn{5}{c}{$16\times 8$} \\
Max turns & 15 & 15 & 15 & 50 & 50 \\
Max tokens per turn & 4{,}096 & 512 & 512 & 512 & 512 \\
History in context & full & last 2 turns & full & last 2 turns & full \\
Max context length & 16{,}384 & 12{,}288 & 16{,}384 & 4{,}096 & 16{,}384 \\
Learning rate & $5\times 10^{-7}$ & $1\times 10^{-6}$ & $5\times 10^{-7}$ & $1\times 10^{-6}$ & $5\times 10^{-7}$ \\
Mini-batch & 8 & 64 & 8 & 256 & 8 \\
KL loss coefficient & \multicolumn{5}{c}{0.01 (low-variance KL)} \\
Entropy coefficient & \multicolumn{5}{c}{0.001} \\
Training steps & 150 & 150 & 150 & 200 & 200 \\
Sampled token IDs & on / off & \multicolumn{4}{c}{on} \\
\bottomrule
\end{tabular}
\end{table*}

\section{More Experiments}\label{app:more_experiments}
\begin{figure*}[t]
\centering
\includegraphics[width=\linewidth]{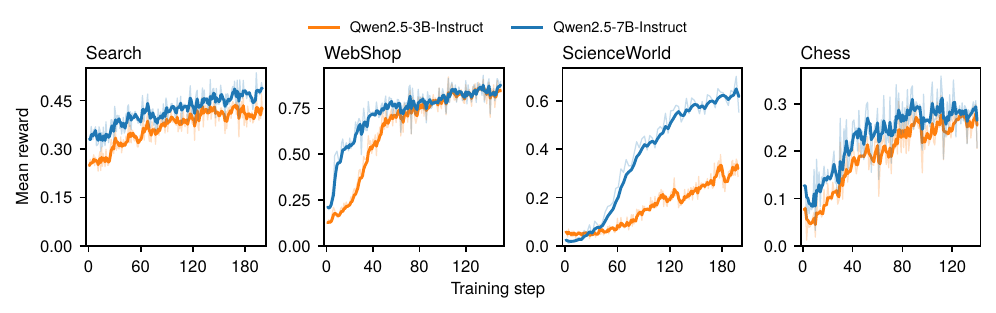}
\caption{Training reward of Qwen2.5-3B-Instruct and Qwen2.5-7B-Instruct on four tasks. Bold lines are exponential moving averages; faint lines are per-step values.}
\label{fig:experiments_envs}
\end{figure*}

\begin{table}[t]
    \centering
    \resizebox{\linewidth}{!}{
    \begin{tabular}{l cc cc cc cc cc cc}
    \toprule
    \multirow{2}{*}{Model} & \multicolumn{2}{c}{Dev} & \multicolumn{2}{c}{Creative} & \multicolumn{2}{c}{CAD} & \multicolumn{2}{c}{Sci} & \multicolumn{2}{c}{Office} & \multicolumn{2}{c}{OS} \\
     & Text & Icon & Text & Icon & Text & Icon & Text & Icon & Text & Icon & Text & Icon \\
    \midrule
QwenVL2.5-3B & 16.2 & 1.4 & 23.3 & 1.4 & 10.2 & 4.7 & 38.2 & 6.4 & 24.3 & 3.8 & 15.0 & 1.1 \\
QwenVL2.5-3B w/ GRPO & 42.9 & 2.1 & 37.4 & 7.7 & 17.8 & 1.6 & 49.3 & 9.1 & 55.4 & 20.8 & 39.3 & 7.9 \\
QwenVL2.5-7B & 33.1 & 2.1 & 23.7 & 3.5 & 12.2 & 6.3 & 36.8 & 7.3 & 37.8 & 7.5 & 30.8 & 6.9 \\
QwenVL2.5-7B w/ GRPO & 46.8 & 3.4 & 39.7 & 8.3 & 19.6 & 5.1 & 52.8 & 10.7 & 58.6 & 22.7 & 42.8 & 9.3 \\
    \bottomrule
    \end{tabular}}
    \caption{Performance of ScreenSpot-Pro GUI grounding.}
    \label{tab:gui_results}
\end{table}

We conduct experiments on the same machines described in \Cref{sec:experiments} for Qwen2.5-3B and 7B.
As shown in \Cref{fig:experiments_envs}, both the 3B and 7B models can be effectively trained with agentic RL, showing clear reward improvements. In general, the 7B model achieves better performance and more significant gains compared to the 3B model.

We also support multi-modal agents, and train Qwen2.5-VL on GUI tasks. To construct our training pipeline, we utilize the GUI-R1 \citep{luo2025gui} dataset. Furthermore, during the reinforcement learning phase, our reward calculation mechanism strictly follows the protocol established in UI-R1 \citep{lu2025ui}. \Cref{tab:gui_results} shows the performance on ScreenSpot-Pro \citep{li2025screenspot}, significantly improving text-grounding capabilities across various domains despite the persisting challenge of purely visual icon grounding.

\section{Tool Implementations}\label{app:tool}
\paragraph{Search} We use the same data and model as in Search-R1 \citep{jin2025search}, and search asynchronously. We randomly sampled examples from HotpotQA \citep{yang-etal-2018-hotpotqa} and Natural Questions \citep{kwiatkowski2019natural} training set as the training data, and directly use the exact match score of the final response and the gold-standard answer as the reward.

\paragraph{ALFWorld} We integrate ALFWorld \citep{shridhar2021alfworld} as an environment service: the stateful simulator runs in an isolated enroot container behind a FastAPI server, and our framework interacts with it through an asynchronous HTTP client. The reward is outcome-based: the reward function queries the server, which checks whether the task's final goal conditions are met and returns a sparse reward ($1.0$ for success, $0.0$ otherwise).

\paragraph{WebShop} We integrate the text-only WebShop environment \citep{yao2022webshop} using the same containerized FastAPI server and asynchronous HTTP client. We adopt WebShop's standard reward, which programmatically matches the attributes, types, options, and price of the purchased product against the target. The environment description is included in the tool definitions and system prompt; the agent navigates with two actions, \tool{search} and \tool{click}, and invalid tool calls decrease as training progresses.

\paragraph{ScienceWorld} We wrap ScienceWorld \citep{wang2022scienceworld} as a containerized FastAPI server accessed via an asynchronous HTTP client. Each task comprises 2--15 sub-goals that yield partial rewards, with full completion earning the maximum reward of $1$; in our multi-turn setting we take the task reward to be the highest reward reached along the trajectory. ScienceWorld has a very large action space, but unlike ALFWorld we list the admissible commands only in the system prompt and tool definition. Despite occasional invalid actions, the agent still issues valid commands and achieves higher average reward as training progresses.

\paragraph{Shell and file tools} The OS tools used for SWE-style tasks (\Cref{sec:prebuilt_tools_tasks}) all operate on one container per rollout. Each call acquires the container through the context under the rollout identifier, so the shell and file tools of a rollout share the same container and file system across all turns, and the container is released when the rollout ends. The container image comes from the task (e.g., the R2E-Gym image of the repository), and commands run in the task's working directory (\texttt{/testbed} by default). Every tool returns at most 10{,}000 characters to the model; longer outputs are truncated.

\texttt{run\_shell\_command(cmd)} runs one command with \texttt{bash} inside the container, with a 120-second timeout and a 4\,GB memory limit, and returns its output, or a short error message if the command times out or exceeds the memory limit. Before execution, a command filter rejects package installation (\texttt{pip}, \texttt{conda}, \texttt{apt}, \texttt{yum}, \texttt{dnf}, \texttt{npm}, \texttt{yarn}), downloads (\texttt{git clone}, \texttt{git lfs pull}, \texttt{wget}, \texttt{curl} downloads), and destructive or resource-exhausting commands (\texttt{rm -rf /}, fork bombs, \texttt{dd} to devices, \texttt{mkfs}, stress tests), returning the reason instead of running the command; installing \texttt{pytest} and an editable install of the repository remain allowed.

The file tools are implemented by a small Python file manager that is copied into the container the first time a file tool runs in a rollout and is invoked with a JSON payload, so they only require a Python interpreter in the image. All paths are resolved relative to the working directory, and paths that escape it are rejected. \Cref{tab:file_tools} lists the tools.

\begin{table}[t]
\centering
\small
\caption{File tools available to SWE agents. All tools run inside the rollout's container.}
\label{tab:file_tools}
\begin{tabular}{>{\raggedright\arraybackslash}p{0.17\linewidth}>{\raggedright\arraybackslash}p{0.25\linewidth}>{\raggedright\arraybackslash}p{0.5\linewidth}}
\toprule
Tool & Arguments & Behavior \\
\midrule
\texttt{read\_file} & \texttt{path}, \texttt{start\_line}, \texttt{end\_line} & Returns the file with line numbers, optionally restricted to a line range. \\
\texttt{create\_file} & \texttt{path}, \texttt{content} & Creates a new file; fails if the path already exists. \\
\texttt{edit\_file} & \texttt{path}, \texttt{search\_block}, \texttt{replace\_block} & Replaces an exact block of text; fails if the block is not found or matches more than one location. The previous content is saved for undo. \\
\texttt{undo\_edit} & \texttt{path} & Restores the file to its content before the last edit, using a per-file snapshot stack stored in a hidden directory of the workspace. \\
\texttt{list\_files} & \texttt{path}, \texttt{max\_depth} & Lists files recursively (default depth 3), skipping hidden, cache, and build directories and binary files. \\
\texttt{grep\_search} & \texttt{pattern}, \texttt{path}, \texttt{include} & Searches for a regular expression with \texttt{grep -rn} (with a Python fallback), optionally filtered by a file glob; returns at most 1{,}000 matches. \\
\texttt{run\_python} & \texttt{path}, \texttt{timeout} & Runs a Python script from the working directory with a timeout (default 60 seconds) and returns its standard output and error. \\
\bottomrule
\end{tabular}
\end{table}

\paragraph{Chess} Chess puzzles are implemented as a \emph{local} (in-process) environment resource rather than a container; each environment holds board state through the \texttt{python-chess} library and calls the \textbf{Stockfish} engine for position evaluation. Puzzles are drawn from the Lichess database, each specifying a FEN starting position and a sequence of UCI solution moves; the environment auto-plays the opponent's first ``setup'' move, and the agent must find all remaining moves.

Three tools share the chess resource acquired for each rollout: \tool{chess\_get\_state} returns the FEN, an ASCII board, side to move, legal-move count, game status, and the moves played so far; \tool{chess\_get\_legal\_moves} lists all legal moves in \texttt{"uci (san)"} format; and \tool{chess\_move} validates a UCI/SAN move, checks it against the solution, and auto-plays the opponent's reply when correct. The reward function acquires the same resource and computes one of two signals. The \textbf{dense} reward is $r = 0.6\,r_{\text{solve}} + 0.4\,r_{\text{quality}}$, where $r_{\text{solve}}$ is $1.0$ for a full solve or partial credit up to $0.5$ for the fraction of correct moves, and $r_{\text{quality}}$ scores each move against Stockfish ($100$ for the engine's best move, else centipawn loss normalized to $0$--$100$ with $-300$\,cp $\rightarrow 0$), averaged and scaled to $[0,1]$. The \textbf{binary} reward is $1.0$ if solved and $0.0$ otherwise.

\end{document}